\documentclass[10pt]{article} 
\usepackage[preprint]{tmlr}

\usepackage{amsmath,amsfonts,bm}

\def\eqref#1{equation~\ref{#1}}

\def\1{\bm{1}}

\DeclareMathAlphabet{\mathsfit}{\encodingdefault}{\sfdefault}{m}{sl}
\SetMathAlphabet{\mathsfit}{bold}{\encodingdefault}{\sfdefault}{bx}{n}

\usepackage{graphicx}   
\usepackage{booktabs}   
\usepackage{subcaption} 

\usepackage{hyperref}
\usepackage{url}
\usepackage{xcolor}

\title{Let It Glow: Controllable Emissification of Non-Emissive Textured 3D Assets with Video Diffusion Priors}

\author{%
  \name Zhiyuan Zhang\textsuperscript{1,2\textdagger}\quad
  Linjun Li\textsuperscript{2}\quad
  Zijian Zhou\textsuperscript{3}\quad
  Haobo Yuan\textsuperscript{1}\quad
  Tiancheng Shen\textsuperscript{1}\quad
  Long Chen\textsuperscript{2}\quad
  Yichen Gong\textsuperscript{2\textdaggerdbl}\quad
  Ming-Hsuan Yang\textsuperscript{1}\thanks{%
    Corresponding author: \nolinkurl{mhyang@ucmerced.edu}\quad
    \textsuperscript{\textdagger}\,\nolinkurl{yx11wng@163.com}\quad
    \textsuperscript{\textdaggerdbl}\,\nolinkurl{yichen.gong@acrab.ai}%
  }\\[3pt]
  {\normalfont\small
    \textsuperscript{1}\,University of California, Merced\quad
    \textsuperscript{2}\,Acrab AI\quad
    \textsuperscript{3}\,King's College London}%
}

\def\month{MM}  
\def\year{YYYY} 
\def\openreview{\url{https://openreview.net/forum?id=XXXX}} 

\begin{document}

\maketitle

\begin{center}
  \includegraphics[width=\textwidth]{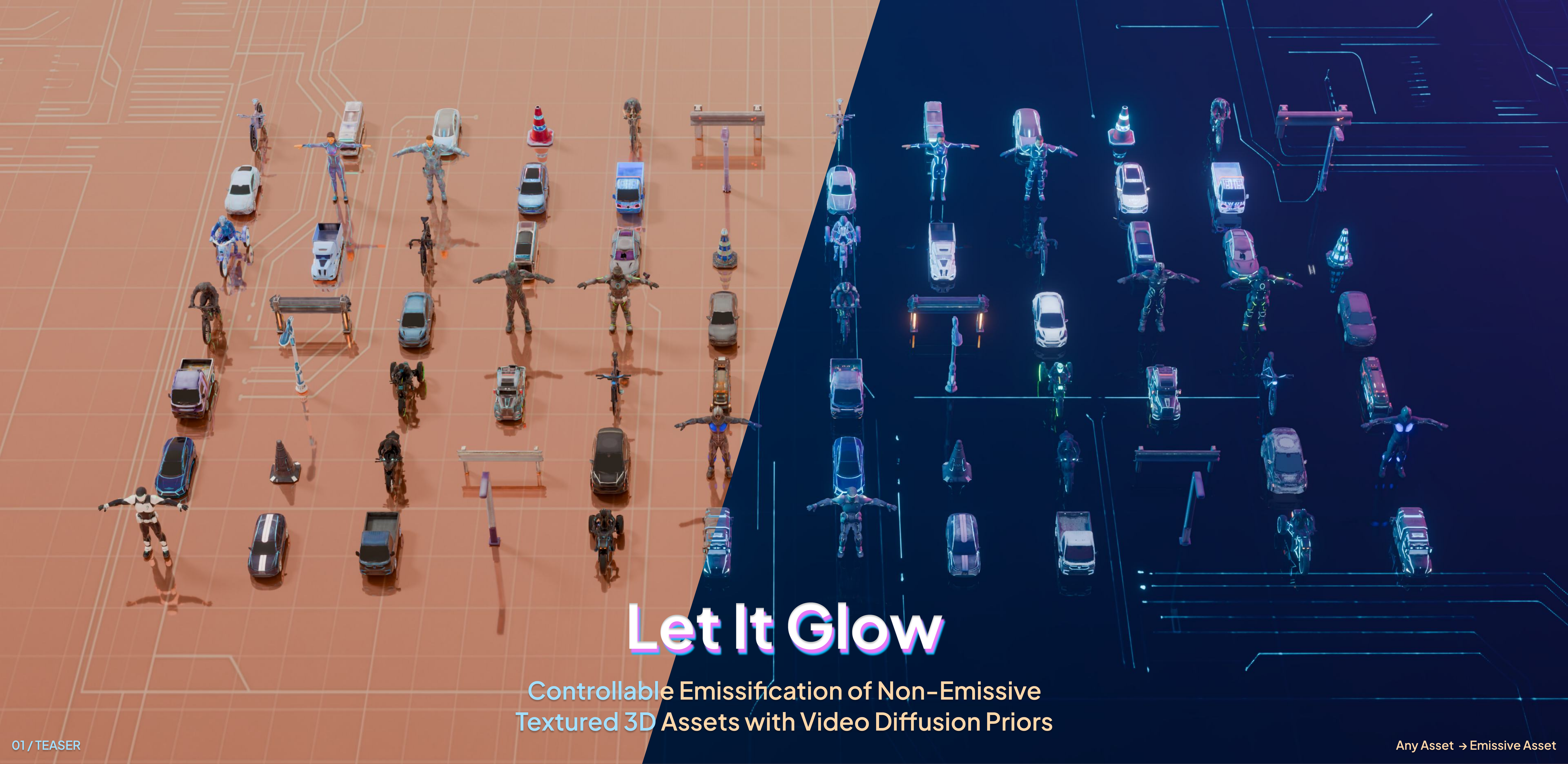}
  \captionof{figure}{\textbf{Controllable emissification of textured 3D assets.} Given a textured 3D object without emissive materials and user guidance in the form of text, a sketch, or both, our method adds emission textures to the specified regions without modifying the geometry or the existing PBR textures. One half of the image renders the objects with all materials enabled; the other half renders the emission channel alone, isolating the generated glow.}
  \label{fig:teaser}
\end{center}

\begin{abstract}

Prior work generates emissive materials for untextured meshes, whereas artists often need to add emission to assets whose materials are already authored. We introduce \emph{non-emissive textured 3D asset emissification}, a task that adds controllable emission to an existing asset while preserving its geometry and all non-emissive PBR textures.
Given text, a sketch, or both, an image editing model produces a reference-aligned emission control image that specifies the intended emitting regions and colors. Our core model, Emissify3D, adapts a pretrained image-to-video diffusion model to generate multi-view consistent emission maps jointly guided by this control image, reference image, and target-view normal maps. A vision-language model then judges the rendered brightness against the user's intent and drives iterative adjustment of a single emission strength while the texture stays fixed.
We train Emissify3D on data constructed from Objaverse-Emission with automatic text annotation and overexposure filtering. On 200 held-out objects, Emissify3D outperforms a general-purpose image editor and an emission generation baseline, and ablations confirm the contribution of each architectural adaptation.

\end{abstract}
    
\section{Introduction}

Textured 3D assets are the building blocks of video games, film, animation, product visualization, and AR/VR applications. In practice, many production assets already come with carefully authored non-emissive PBR materials, including albedo, metallic, roughness, and normal maps. Yet the visual identity of such an asset is often defined less by how it reflects light than by how it emits light. Neon signage, sci-fi armor seams, enchanted runes, and holographic interfaces are perceived primarily through their emission rather than their geometry or reflectance. Emissive regions dictate appearance in low-light environments and establish the focal hierarchy that directs viewer attention. Because self-illumination has a visual impact out of proportion to its surface area, a localized emission edit can change an asset's appearance far more than a comparable edit to its albedo or geometry.

\noindent \textbf{The Omitted Emission Channel.} Despite rapid progress in 3D texture and PBR material generation~\cite{chen2023text2tex, yu2024texgen, zeng2024paint3d, huang2025material, hunyuan3d2025hunyuan3d}, this channel has received remarkably little attention. The dominant paradigm takes an untextured mesh and synthesizes the reflectance stack, typically albedo, metallic, and roughness, so that the asset renders plausibly under arbitrary illumination. These maps describe only how a surface responds to incident light, and consequently cannot express a surface that emits light on its own. The recent EmissionGen~\cite{zhang2026towards} generates emission textures for untextured meshes from a reference image of an emissive object and is, to our knowledge, the only prior work that treats emission as a first-class output. Because it is formulated for bare geometry, it does not address assets whose materials have already been authored, nor does it offer explicit control over where emission appears or what color it takes.

\noindent \textbf{An Unaddressed Editing Setting.} The need that arises in practice is different from generation. Artists work with assets that are already textured, and wish to add localized self-illumination, such as neon trims, glowing runes, sci-fi panels, or luminous signage, while preserving the existing materials exactly. This setting is fundamentally more constrained than generating a complete material stack from scratch: the new emission must be compatible with the asset's original reflectance, geometry, and artistic style rather than replacing them, and its spatial extent and color must follow what the artist actually intends. We are not aware of a method that addresses this setting. Approaches that regenerate the full set of texture maps overwrite the original PBR materials, while methods formulated for bare meshes presuppose that no materials exist to preserve. Artists therefore still rely on manual workflows: inspecting UV layouts, painting emission masks island by island, tuning emission strength, and repeatedly re-rendering the asset to check whether the glow matches the original material design.

\noindent \textbf{Our work.} To address this limitation, we introduce \emph{non-emissive textured 3D asset emissification}, a novel task that adds emissive materials to an existing textured 3D asset whose materials contain no emission, following a text prompt or coarse visual guidance or both. Unlike PBR generation, the task explicitly preserves the input geometry and all non-emissive PBR maps, and produces only a UV-space emission map together with a calibrated emission strength. This formulation matches a common production need: enabling artists to create controllable glowing effects without damaging or re-authoring the asset's original material stack. To the best of our knowledge, this task has not been formulated before.

\noindent \textbf{Key Challenges.} Emissification is not simply PBR texturing with one more channel; adding emission to an already-textured asset raises four coupled difficulties. First, text guidance may be spatially underspecified: a prompt such as ``glowing blue accents'' describes the desired appearance without specifying where the glow belongs, admitting many valid placements and calling for a control signal that constrains emission location and color without over-constraining global appearance. Second, the emission must attach to semantically meaningful structures such as panel seams, window frames, and engravings, hold across all viewpoints, and stay stylistically compatible with the artist's existing reflectance; view-inconsistent predictions survive back-projection as seams and blurred boundaries in UV space. Third, brightness is governed by a scalar emission strength decoupled from the texture map. Too high overexposes the render, too low leaves the effect imperceptible, and the right magnitude is subjective, asset-dependent, and observable only in the rendered result. Finally, supervision must be curated rather than collected: existing emissive assets carry no text describing their emissive appearance, and a portion of their renderings is already overexposed.

Our method first uses an image editing model to convert user guidance into an emission control image aligned with the asset's reference rendering. The control image specifies the desired emission regions and colors. Three targeted adaptations repurpose an image-to-video diffusion model to generate view-consistent emission maps jointly conditioned on the control image, reference rendering, target-view normal maps, and a text prompt. We then bake the maps into the asset's UV space. With the resulting texture fixed, we iteratively adjust the emission strength using feedback from a vision-language model that assesses rendered brightness against the user's preference.

To train the model, we build an emissification training set from the 40K assets of Objaverse-Emission~\cite{zhang2026towards}. Each sample is paired with an automatically generated caption describing the object and its emissive appearance, and overexposed renderings are filtered out.

Our contributions are as follows:
\begin{itemize}
    \item We define \emph{non-emissive textured 3D asset emissification}, a novel preservation-oriented 3D material editing task that adds controllable emission to existing non-emissive textured assets while keeping geometry and all non-emissive PBR materials unchanged.
    \item We propose a pipeline that converts free-form user intent into a canonical, reference-aligned emission control image, and that tunes emission strength with VLM feedback without changing the generated texture.
    \item We repurpose an image-to-video diffusion model for multi-view emission map generation through three adaptations (multi-image channel conditioning, a temporal RoPE stride, and frame replication) and ablate each of them.
    \item We construct a 3D emissification training set with automatic captioning and overexposure filtering, and show on 200 held-out objects that Emissify3D predicts target-view emission maps more faithfully than other baselines.
\end{itemize}

\section{Related Work}
\label{sec:related}

\noindent \textbf{Texture Generation and Editing.}
Automatic texture synthesis for 3D assets has advanced rapidly in recent years. Early approaches transfer 2D diffusion priors onto mesh surfaces through score distillation~\cite{poole2022dreamfusion} or iterative view-by-view inpainting~\cite{richardson2023texture,chen2023text2tex}, whereas recent methods synthesize multi-view images and fuse them into unified UV textures, yielding superior cross-view consistency and inference efficiency~\cite{li2025step1x,feng2025romantex}.

A prominent direction targets physically based rendering (PBR) materials rather than baked RGB textures~\cite{hunyuan3d2025hunyuan3d,he2025materialmvp,huang2025material,wang2024boosting}. These methods disentangle albedo, metallic, and roughness maps, enabling realistic rendering under arbitrary lighting in standard graphics engines, and have become the dominant paradigm for texturing untextured geometry.

These methods model surface \emph{reflectance} only and omit the \emph{intrinsic light emission} needed for glowing runes, neon signage, electronic displays, magical effects, and sci-fi panels.

The most closely related work is EmissionGen~\cite{zhang2026towards}, which introduced emissive material generation for 3D objects together with the Objaverse-Emission dataset that we build on. EmissionGen is conditioned on a reference image and targets untextured meshes; it provides no explicit \emph{spatial} or \emph{chromatic} control and does not aim to preserve existing materials. Our work instead edits already-textured assets and specifies where and in which color emission should appear.

\noindent \textbf{Instruction-Driven 2D and 3D Editing.}
Instruction- and condition-driven image editors~\cite{brooks2023instructpix2pix,zhang2023adding,nanobanana2025} make it easy to specify local appearance changes in a single view, and several 3D editing methods lift such edits to multiple views by iteratively editing renderings and re-optimizing the 3D representation~\cite{haque2023instruct}. These methods edit the shaded appearance; they do not produce a separate emission channel that remains a light source under new illumination. We use an image editor only to create the reference-aligned control image and train a separate model to generate multi-view emission maps.

\noindent \textbf{Video Generation Models for 3D Tasks.}
Large-scale video generation models~\cite{wan2025wan,kong2024hunyuanvideo} are trained on vast collections of real-world footage depicting objects under continuous camera motion. They implicitly encode rich 3D priors, including occlusion, motion parallax, shading consistency, and viewpoint-dependent appearance evolution~\cite{wan2025wan}. A growing body of work leverages these priors for 3D tasks by framing multi-view sequences as short videos and repurposing pretrained video generation models as multi-view generators~\cite{yuan2025seqtex,munkberg2025videomat,liu2026scaling,voleti2024sv3d}. Unlike multi-view image diffusion models that must learn cross-view consistency from scratch~\cite{hunyuan3d2025hunyuan3d}, video-based formulations inherit temporal attention that naturally transfers into view consistency. Recent analyses indicate that video foundation models encode substantial 3D structure~\cite{huang2026much}, and video-based multi-view generators have been reported to need less task-specific 3D training data than models trained from image backbones. This property is particularly advantageous in our context, given the scarcity of annotated emissive 3D assets, motivating our choice of an image-to-video backbone for multi-view emission map generation.

\noindent \textbf{Tool-Assisted Generation.}
Recent systems use language and vision-language models to coordinate specialized tools for 3D scene and image generation~\cite{ling2026scenethesis,ye2026genclaw}, or to iteratively refine material and lighting parameters in a graphics program from rendered feedback~\cite{huang2024blenderalchemy}. In our setting, an image editor translates user guidance into a reference-aligned emission control image, and a vision-language model assesses rendered brightness after the emission texture has been generated. This feedback adjusts only the scalar emission strength and never regenerates the texture.

\begin{figure*}[!t]
\centering
\includegraphics[width=\textwidth]{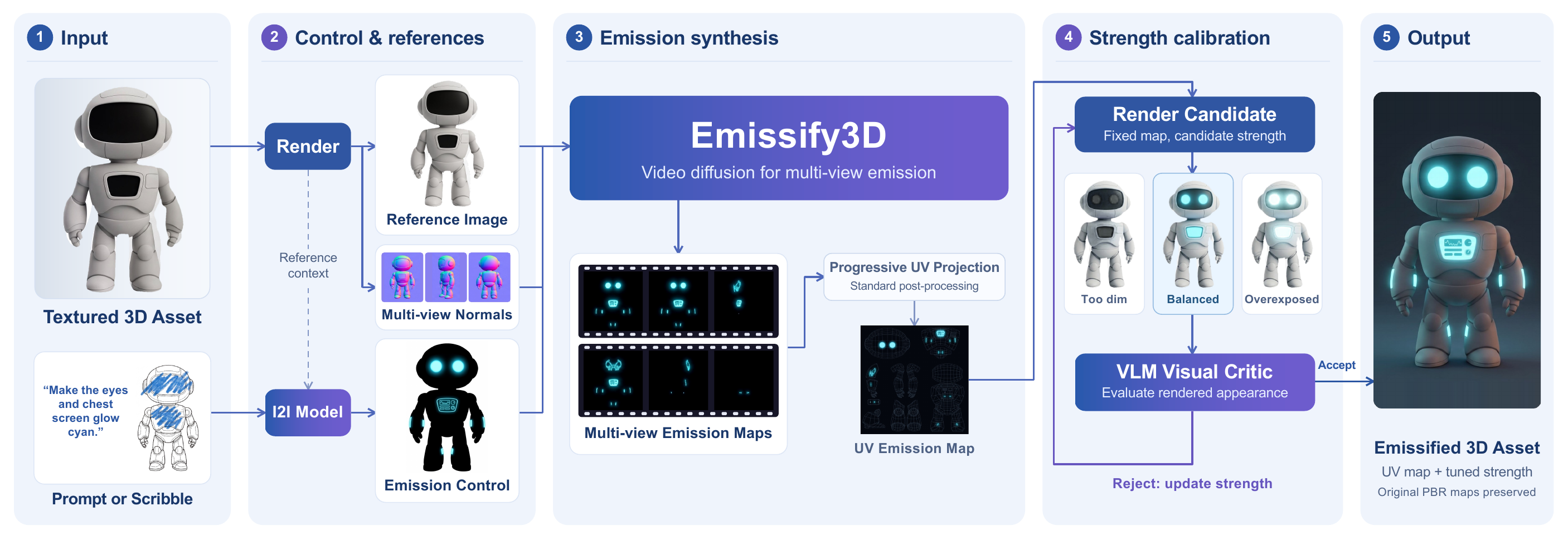}
\caption{\textbf{Emissification pipeline.} An image editing model combines user guidance with the asset's reference rendering to produce an emission control image. Emissify3D generates view-consistent emission maps for the target views. We then back-project the emission maps into UV space. VLM feedback on renders guides iterative adjustment of a global emission strength while the texture remains fixed.}
\label{fig:pipeline}
\end{figure*}

\section{Method}
\label{sec:method}

Our framework turns a user's text or sketch guidance into an explicit emissive material for an existing non-emissive textured 3D asset, and separates \emph{what} emits from \emph{how strongly} it emits. A reference-aligned emission control image specifies the desired emission regions and colors. Guided by this control image, Emissify3D generates emission maps for the target views. We bake generated emission maps into a UV texture and adjust its global emission strength using feedback on rendered brightness while keeping the texture fixed. We first define the task (Sec.~\ref{sec:task}), then describe the pipeline (Sec.~\ref{sec:framework}) and the core multi-view model (Sec.~\ref{sec:core_model}).

\subsection{Task Definition}
\label{sec:task}

Let $\mathcal{O}$ denote an existing non-emissive textured 3D asset and $u$ denote the user's editing guidance, provided as text, a coarse sketch, or both. Our goal is to produce an edited asset $\mathcal{O}_{\mathrm{emit}}$, together with its emission texture $\mathcal{T}_{\mathrm{emit}}$ and emission strength $s_{\mathrm{emit}}$:
\begin{equation}
\begin{aligned}
(\mathcal{T}_{\mathrm{emit}},s_{\mathrm{emit}})
&\mathrel{:=}\mathcal{F}(\mathcal{O},u)\\
\mathcal{O}_{\mathrm{emit}}
&\mathrel{:=}(\mathcal{O};\mathcal{T}_{\mathrm{emit}},s_{\mathrm{emit}})
\end{aligned}
\label{eq:task}
\end{equation}
Here, $\mathcal{F}$ denotes the complete framework that generates $\mathcal{T}_{\mathrm{emit}}$ and $s_{\mathrm{emit}}$. 
The edited asset retains the source geometry and all non-emissive PBR textures. The emission texture encodes the spatial extent and color of emission, while the emission strength scales its intensity at render time.

\begin{figure*}[!t]
\centering
\includegraphics[width=\textwidth]{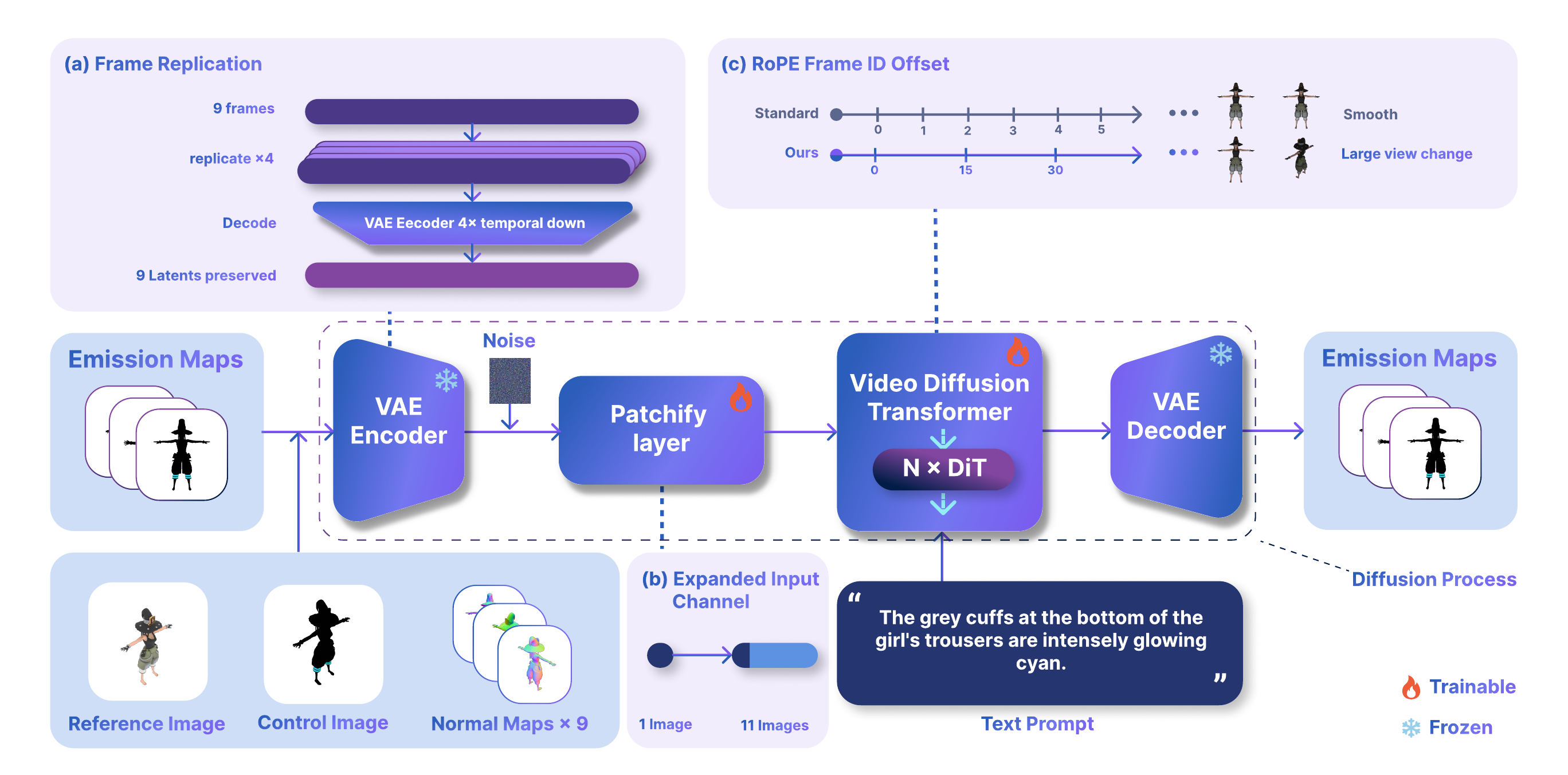}
\caption{\textbf{Emissify3D architecture.} We adapt an image-to-video generation model for multi-view emission map generation. Key modifications: (a) frame replication to counteract VAE temporal compression; (b) an expanded patchify input layer that accepts $2+K$ concatenated conditioning images (reference, control, and $K$ normal maps); and (c) a temporal RoPE stride that accommodates large viewpoint changes between consecutive frames.}
\label{fig:model}
\end{figure*}

\subsection{Emissification Pipeline}
\label{sec:framework}

Figure~\ref{fig:pipeline} shows the full pipeline. We first construct the emission texture through rendering, control generation, multi-view emission map generation, and UV projection:
\begin{equation}
\begin{aligned}
(\mathcal{I}_{\mathrm{ref}},\{\mathcal{N}_k\}_{k=1}^{K})
&=\mathcal{R}(\mathcal{O})\\
(\mathcal{I}_{\mathrm{control}},p)
&=\mathcal{S}_{\mathrm{ctrl}}(u,\mathcal{I}_{\mathrm{ref}})\\
\{\mathcal{E}_k\}_{k=1}^{K}
&=\mathcal{G}_{\theta}(\mathcal{I}_{\mathrm{ref}},
\mathcal{I}_{\mathrm{control}},\{\mathcal{N}_k\}_{k=1}^{K},p)\\
\mathcal{T}_{\mathrm{emit}}
&=\mathcal{P}(\mathcal{O},\{\mathcal{E}_k\}_{k=1}^{K})
\end{aligned}
\label{eq:pipeline}
\end{equation}
Here, the renderer $\mathcal{R}$ produces the full-material reference image $\mathcal{I}_{\mathrm{ref}}$ and the normal maps $\{\mathcal{N}_k\}_{k=1}^{K}$ for $K$ target views. The control generation stage $\mathcal{S}_{\mathrm{ctrl}}$ provides the emission control image $\mathcal{I}_{\mathrm{control}}$ and text conditioning $p$ for Emissify3D, denoted by generator $\mathcal{G}_{\theta}$. We use progressive back-projection~\cite{huang2025material,hunyuan3d2025hunyuan3d} for $\mathcal{P}$ to bake the predicted emission maps $\{\mathcal{E}_k\}_{k=1}^{K}$ into $\mathcal{T}_{\mathrm{emit}}$.

\noindent \textbf{Reference-guided emission control generation.} Text and sketches do not specify emission in the pixel-aligned format expected by Emissify3D: text is spatially underspecified, and a free-hand sketch is only roughly aligned with the asset. We therefore condition an image editing model on the asset's reference rendering and the user's guidance. When the user provides text, we interpret it to construct an editing prompt $p$, which is part of the editor's input and also conditions Emissify3D. The editor outputs $\mathcal{I}_{\mathrm{control}}$ in the reference view, using the target colors for intended emitting pixels and black elsewhere. This format matches the reference-view emission passes used as control images during training.

\noindent \textbf{VLM-guided emission strength adjustment.} Once the UV texture is baked, we adjust its emission strength based on how bright the asset appears after rendering. At iteration $t$,
\begin{equation}
\begin{aligned}
\mathcal{Y}_{t}
&=\mathcal{R}(\mathcal{O};\mathcal{T}_{\mathrm{emit}},s_t)\\
q_t&=\mathcal{V}(\mathcal{Y}_{t},b)\\
s_{t+1}&=\mathcal{U}(s_t,q_t)
\end{aligned}
\label{eq:strength_adjustment}
\end{equation}
The renderer produces $\mathcal{Y}_t$ using the fixed emission texture at strength $s_t$. The VLM $\mathcal{V}$ assesses the rendered brightness, taking into account the user's brightness description $b$ when provided, and returns one of three judgments: \emph{too dim}, \emph{appropriate}, or \emph{too bright}. The rule $\mathcal{U}$ increases, retains, or decreases the strength, respectively, and the loop terminates when $\mathcal{V}$ returns \emph{appropriate} or a maximum number of iterations is reached. Across iterations only the scalar changes; the emission texture and its spatial and chromatic content remain fixed.

\begin{figure*}[!t]
\centering
\includegraphics[width=\textwidth]{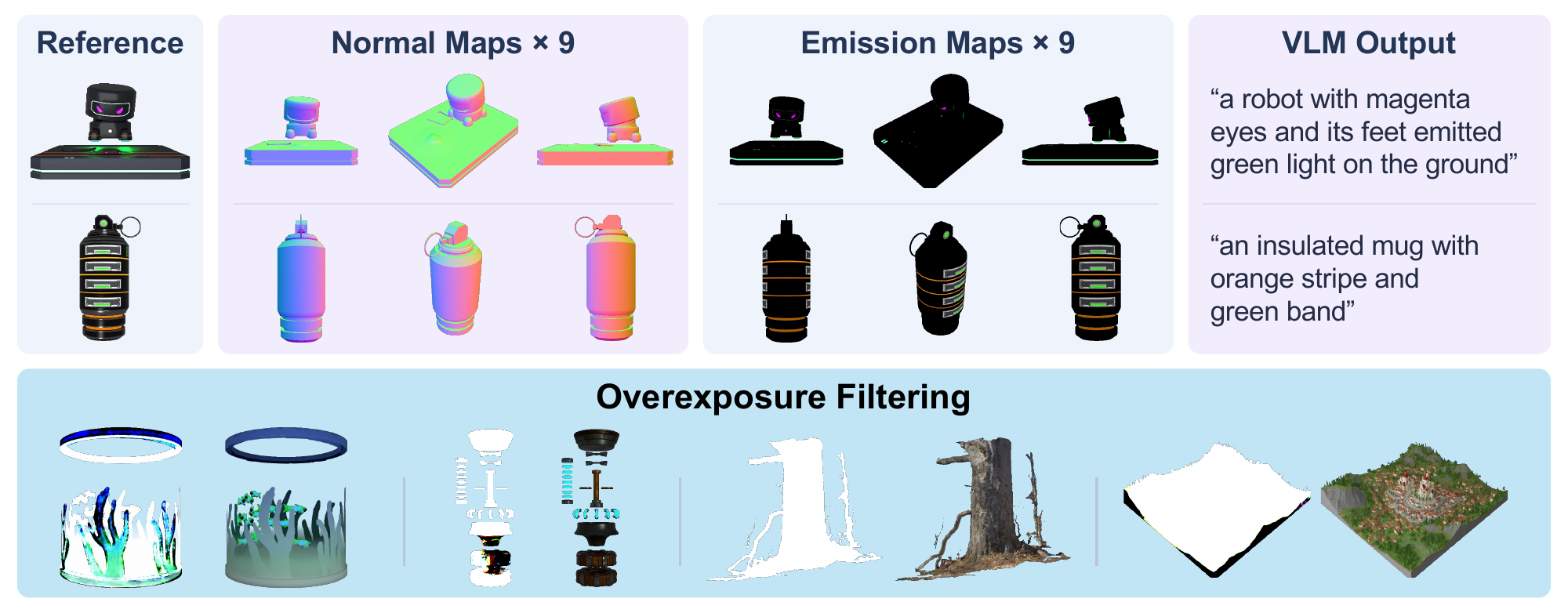}
\caption{\textbf{Training data examples.} \textbf{Top}: Each example shows a full-material reference rendering, paired normal and ground-truth emission maps for three of the nine target views, and a VLM-generated caption used for text conditioning. \textbf{Bottom}: Each pair shows the same asset rendered at two emission strengths. We discard the overexposed rendering on the left and retain the valid one on the right.}
\label{fig:data}
\end{figure*}

\subsection{Core Emissification Model}
\label{sec:core_model}

The core of our approach is Emissify3D, which adapts a pretrained video diffusion model for multi-view emission map generation. We use a video backbone because video generation models are trained on large amounts of footage of objects seen from moving cameras and therefore encode useful priors for cross-view consistency (Sec.~\ref{sec:related}), which is useful given the scarcity of annotated emissive 3D assets.

\noindent \textbf{Model Formulation.} Emissify3D generates multi-view emission maps, conditioned on a reference rendering, an emission control image, target-view normal maps, and a text prompt:
\begin{equation}
\text{Emissify3D}(\mathcal{I}_{\mathrm{ref}}, \mathcal{I}_{\mathrm{control}}, \{\mathcal{N}_k\}_{k=1}^{K},p) = \{\mathcal{E}_k\}_{k=1}^{K}
\end{equation}
where the two reference-view images play distinct roles: $\mathcal{I}_{\mathrm{ref}}$ captures the source appearance with all materials enabled, while $\mathcal{I}_{\mathrm{control}}$ specifies the target emission regions and colors. At inference the source asset is non-emissive, so $\mathcal{I}_{\mathrm{ref}}$ contains no glow. The prompt $p$ provides semantic context, while each normal map $\mathcal{N}_k$ provides geometric conditioning for output view $k$. We sample the $K$ target viewpoints uniformly in azimuth around the object and order them along the camera trajectory to exploit the video generation model's temporal coherence.

\noindent \textbf{Input Encoding.} A standard image-to-video (I2V) model accepts a single conditioning image: the image is encoded by the VAE, the resulting latent is concatenated channel-wise with the noise latent, and the result is fed to a patchify layer. We instead condition on $2+K$ images (the reference image, the control image, and $K$ normal maps). We encode each with the VAE, concatenate all conditioning latents with the noise latent along the channel dimension, and widen the input channels of the patchify layer accordingly. All other pretrained transformer blocks are kept unchanged.

\noindent \textbf{Temporal Adaptation for Multi-view Generation.} Video generation models produce temporally smooth sequences in which consecutive frames differ only slightly, whereas consecutive views in our sequence differ by a large rotation ($360^\circ/K$ in azimuth). We therefore rescale the temporal Rotary Position Embedding (RoPE): the $t$-th latent frame receives temporal position $\sigma t$ instead of $t$, with stride $\sigma=15$. The stride is chosen so that the positional gap between consecutive views roughly matches the number of frames over which a comparable camera rotation unfolds in ordinary video, encouraging larger inter-frame changes.

\noindent \textbf{Handling VAE Temporal Compression.} The video VAE compresses time by a factor of $4$ except for the first frame, mapping $1+4n$ input frames to $1+n$ latent frames~\cite{wan2025wan}. Encoding our $K$ distinct views directly would merge groups of four different viewpoints into a single latent frame. We therefore repeat every view except the first four times, giving $1+4(K-1)$ frames ($33$ for $K=9$) that the encoder maps to exactly $K$ latent frames, one per view. Replication is applied during both training and inference.

\section{Experiments}
\label{sec:exp}

\subsection{Experimental Setup}
\label{sec:exp_setup}

\noindent \textbf{Training Data.} We construct the training set from Objaverse-Emission~\cite{zhang2026towards}, which provides 40K 3D assets with emission maps, normal maps, and full-material reference renderings. For each sample, $\mathcal{I}_{\mathrm{ref}}$ is the full-material rendering of the original asset, $\mathcal{I}_{\mathrm{control}}$ is the ground-truth emission pass from the same reference view, and the target-view ground-truth emission maps provide supervision together with their normal maps. We vary environment illumination when rendering $\mathcal{I}_{\mathrm{ref}}$, while $\mathcal{I}_{\mathrm{control}}$ and the target emission maps are derived from the same illumination-independent ground-truth emission texture~\cite{hunyuan3d2025hunyuan3d}. Emission strength also varies across samples; within each sample, the same strength is used for $\mathcal{I}_{\mathrm{ref}}$ and the target emission maps~\cite{zhang2026towards}.

We use Qwen3-VL~\cite{bai2025qwen3} to generate template-style prompts for text conditioning during training (examples in Fig.~\ref{fig:data}). These prompts provide coarse semantic context, while $\mathcal{I}_{\mathrm{control}}$ specifies the emission regions and colors. The source dataset also contains renderings that become overexposed at excessive emission strengths. A VLM inspects these samples and removes only the overexposed renderings, retaining valid renderings of the same asset at other strengths.

\noindent \textbf{Implementation Details.} We build our core emissification model Emissify3D on Wan2.1-I2V-14B-480P~\cite{wan2025wan}, a large-scale image-to-video diffusion transformer. Rather than full fine-tuning, we adopt LoRA~\cite{hu2021lora} with rank 128, which preserves the pretrained video priors while adapting the model to the emissification task with a tractable number of trainable parameters. We use $K=9$ target views for each object. All images are rendered at $512\times512$ resolution. We set the temporal RoPE stride to $15$, and match the VAE's temporal compression factor of $4$ by expanding every frame except the first to four copies. We widen the input channel of the patchify layer to accommodate eleven concatenated visual inputs: one reference image, one emission control image, and nine normal maps. The expanded patchify layer is trained jointly with the LoRA weights, since its input dimension differs from the pretrained checkpoint and cannot be inherited. We optimize with AdamW at a learning rate of $1\times10^{-4}$ for approximately $20$K steps on $24$ H20 GPUs.

We use Nano Banana (Gemini 2.5 Flash Image)~\cite{nanobanana2025} as the image editor in the control generation stage $\mathcal{S}_{\mathrm{ctrl}}$, Qwen3-VL~\cite{bai2025qwen3} as the brightness evaluator $\mathcal{V}$, and Blender's Cycles engine as the renderer $\mathcal{R}$.

\noindent \textbf{Training Objective.} We train Emissify3D with the standard flow-matching velocity-prediction objective of the backbone~\cite{wan2025wan}: given a noised latent of the target multi-view emission maps, the model predicts the flow velocity conditioned on the reference image, emission control image, target-view normal maps, and text prompt.

\noindent \textbf{Benchmark.} We evaluate on $200$ objects from Objaverse-Emission held out from training. For each object, we render a reference image and normal maps for nine target views following the training protocol. We use the ground-truth emission map rendered from the reference view as the emission control image. This setup evaluates multi-view emission map generation independently of control generation, UV baking, and strength adjustment.

\noindent \textbf{Metrics.} PSNR, SSIM~\cite{wang2004image}, and LPIPS~\cite{zhang2018unreasonable} measure pixel-level and perceptual fidelity of the generated multi-view emission maps against ground truth. FID~\cite{heusel2017gans} and CLIP-FID~\cite{kynkaanniemi2023role} compare the distributions of generated and ground-truth emission maps using Inception and CLIP features, respectively. CLIP-I is the cosine similarity between the CLIP~\cite{radford2021learning} image embeddings of the generated emission maps and the emission control image. Because most pixels of an emission map are black, full-image PSNR and SSIM are dominated by correctly predicted background.

\subsection{Quantitative Evaluation}
\label{sec:exp_quant}

We compare Emissify3D with EmissionGen~\cite{zhang2026towards} and Nano Banana~\cite{nanobanana2025} on the $200$ held-out objects. All methods are evaluated against the same ground-truth emission maps. This comparison evaluates how well each method generates multi-view emission maps from a single-view specification of emitting regions and colors, not how well the control image is generated from user input.

\noindent \textbf{EmissionGen.} EmissionGen is the closest baseline to our task, but it generates emissive materials for untextured meshes rather than editing textured assets. It conditions on a full-material RGB image of an emissive object and uses normal and position maps as geometry control. To match its reference format, we use image editing model~\cite{nanobanana2025} to apply the regions and colors specified by the ground-truth emission control image to the same non-emissive reference rendering used by Emissify3D. We render the normal and position maps from the ground-truth mesh using the same nine target camera poses as Emissify3D. Metrics are evaluated the predicted emission maps generated by EmissionGen.

\noindent \textbf{Nano Banana.} As a standalone baseline, Nano Banana receives the same visual inputs as Emissify3D, which are the non-emissive reference rendering, the ground-truth emission control image, and the target-view normal maps. We apply it independently to each target view.

\begin{table}[t]
\centering
\caption{\textbf{Multi-view emission map generation on 200 held-out objects.} All methods receive the ground-truth emission control image from the reference view. Metrics are computed on the generated target-view emission maps.}
\label{tab:emissification_quant}
\renewcommand{\arraystretch}{0.9}
\resizebox{0.65\columnwidth}{!}{
\begin{tabular}{lcccccc}
\toprule[1.2pt]
Method & PSNR${\uparrow}$ & SSIM${\uparrow}$ & LPIPS${\downarrow}$ & CLIP-FID${\downarrow}$ & FID${\downarrow}$ & CLIP-I${\uparrow}$ \\
\midrule
Emissify3D & \textbf{37.10} & \textbf{0.946} & \textbf{0.0839} & \textbf{0.109} & \textbf{42.63} & \textbf{0.799} \\
EmissionGen & 35.83 & 0.875 & 0.1061 & 0.164 & 47.52 & 0.691 \\
Nano Banana & 31.45 & 0.702 & 0.1280 & 0.193 & 58.74 & 0.586 \\
\bottomrule[1.2pt]
\end{tabular}
}
\vspace{-10pt}
\end{table}

\begin{figure}[!htbp]
\centering
\includegraphics[width=\textwidth]{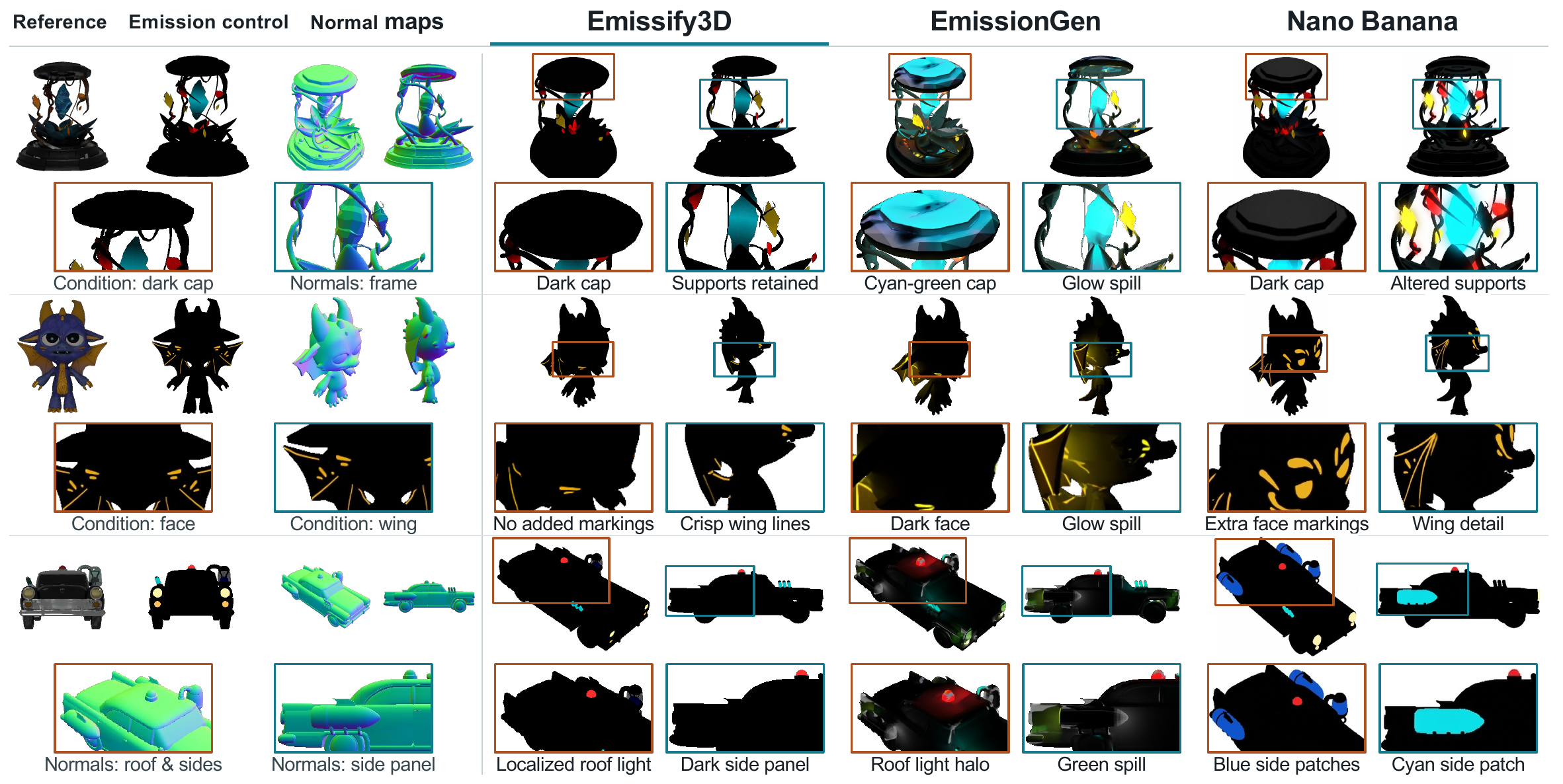}
\caption{\textbf{Qualitative comparison.} Given an input 3D object and an emission control image, we compare Emissify3D with EmissionGen and Nano Banana. Emissify3D follows the control image and keeps the emissive regions consistent across views. EmissionGen adds emission outside the specified regions (e.g., the cyan-green cap and glow spill), and Nano Banana alters or adds markings that vary between views.}
\label{fig:qual_comparison}
\end{figure}

\begin{figure*}[!t]
\centering
\includegraphics[width=\textwidth]{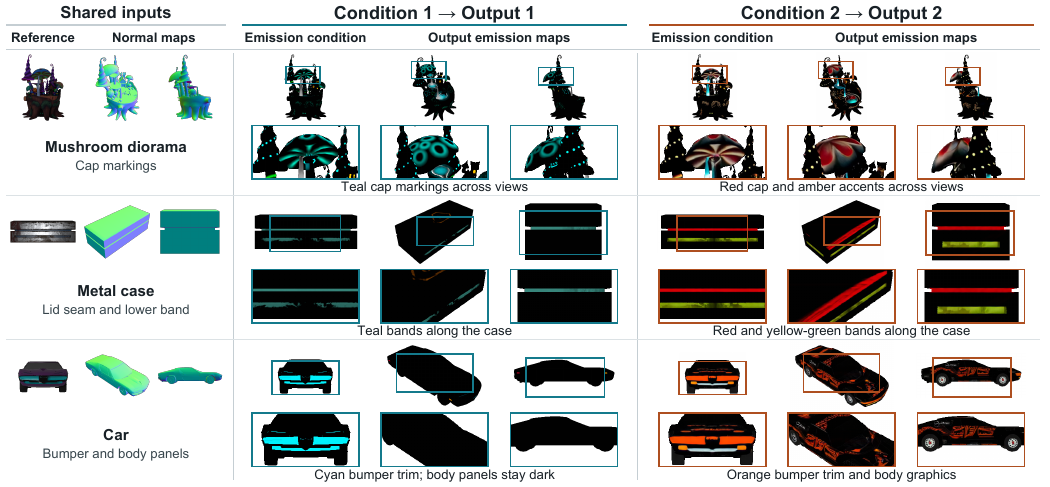}
\caption{\textbf{Controllable emission map generation.} For a fixed input object, we vary only the emission control image and run Emissify3D independently for each. The generated multi-view emission maps follow the regions and colors specified by each control image while remaining consistent across viewpoints, so a single asset can be re-styled by re-specifying the control condition alone.}
\label{fig:editing}
\end{figure*}

\begin{figure*}[!t]
\centering
\includegraphics[width=\textwidth]{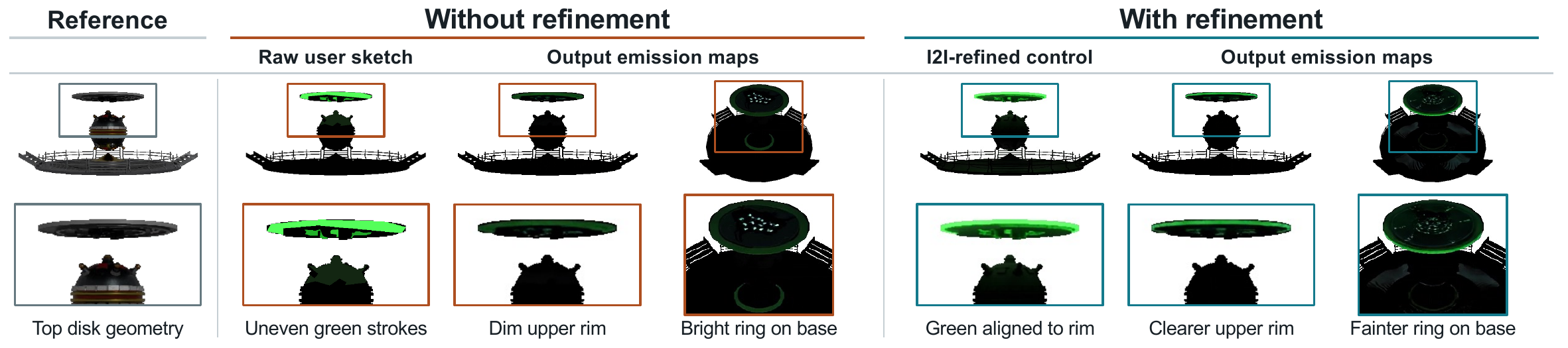}
\caption{\textbf{Effect of emission control generation.} Without image-editor refinement of the user sketch, Emissify3D receives an ambiguous control signal and produces degraded multi-view emission maps with inaccurate regions and inconsistent glow patterns. After refinement, the emission control image becomes semantically aligned with the input asset, enabling Emissify3D to generate appropriate and view-consistent emission maps.}
\label{fig:control_generation_ablation}
\end{figure*}

Emissify3D achieves the best results on all six metrics (Table~\ref{tab:emissification_quant}). Compared with EmissionGen, it improves PSNR from $35.83$ to $37.10$~dB, SSIM from $0.875$ to $0.946$, and CLIP-I from $0.691$ to $0.799$.

Although EmissionGen also models interactions between views, Emissify3D uses cross-frame attention from a pretrained video backbone to generate the target-view emission maps jointly, drawing on learned priors for consistent object appearance under camera view motion. Emissify3D also receives an emission control image that shows only the intended emission. In contrast, EmissionGen must infer the intrinsic emitting regions and colors from a rendered image that mixes emission with surface appearance and lighting. This ambiguity makes it less suited to editing only the emission channel of a textured asset: it can introduce incorrect glare outside the specified regions (Fig.~\ref{fig:qual_comparison}).

Nano Banana edits each view independently, so the position, shape, and color of emitting regions can drift across views, producing seams or blurred boundaries after UV projection. It can also miss specified regions or add emission outside them, reducing fidelity to the control image.

\subsection{Qualitative Comparison and Evaluation}
\label{sec:exp_qual}

\noindent \textbf{Baselines.} Because no existing method targets 3D emissification editing directly, we compare against the closest available alternatives under the protocol of Sec.~\ref{sec:exp_quant}: EmissionGen~\cite{zhang2026towards}, which generates emission for 3D objects, and Nano Banana~\cite{nanobanana2025}, a strong general-purpose image editor that receives the same emission control image as Emissify3D.

\noindent \textbf{Comparison.} Figure~\ref{fig:qual_comparison} presents visual comparisons. Nano Banana's emissive regions drift in position and color across viewpoints and do not follow the patterns of the control image, e.g., by adding face markings or blue side patches that are absent from the specification. Because each view is edited independently, nothing enforces cross-view agreement, and projecting such views into UV space produces seams and blurred emission boundaries. EmissionGen produces plausible glow but places it on regions the control image leaves dark.

Emissify3D produces emission maps that follow the control specification and remain consistent across viewpoints. Because the result is a separate emission texture rather than an edited shaded image, the emissified asset acts as a light source under new illumination and viewpoints.

\noindent \textbf{Custom Emission Control Editing.}
We probe controllability by fixing the reference image and target-view normal maps and varying only the emission control image. As shown in Figure~\ref{fig:editing}, the generated emission follows the specified regions and colors and leaves unmarked areas non-emissive. Emissify3D therefore does not collapse to a single plausible pattern per object; it propagates the control image to unseen viewpoints while keeping emissive regions attached to consistent surface structures.

\noindent \textbf{Scope of the evaluation.} The results above isolates multi-view emission map generation using ground-truth control images. The contribution of the control-generation stage, which converts user guidance into emission control images, is examined in Sec.~\ref{sec:ablation}.

\subsection{Ablation Study}
\label{sec:ablation}

\begin{table}[t]
\centering
\caption{\textbf{Ablation of the Emissify3D architecture} on the 200-object benchmark. $\sigma$ denotes the temporal RoPE stride (full model: $\sigma=15$).}
\label{tab:ablation}
\renewcommand{\arraystretch}{0.9}
\resizebox{0.85\columnwidth}{!}{
\begin{tabular}{lcccccc}
\toprule[1.2pt]
Method & PSNR${\uparrow}$ & SSIM${\uparrow}$ & LPIPS${\downarrow}$ & CLIP-FID${\downarrow}$ & FID${\downarrow}$ & CLIP-I${\uparrow}$ \\
\midrule
Emissify3D (full) & \textbf{37.10} & \textbf{0.946} & \textbf{0.0839} & \textbf{0.109} & \textbf{42.63} & \textbf{0.799} \\
\midrule
No frame replication 	& 32.48 & 0.751 & 0.1207 & 0.180 & 53.49 & 0.614 \\
RoPE stride $\sigma=1$ (original) & 30.76 & 0.695 & 0.1384 & 0.217 & 63.76 & 0.561 \\
RoPE stride $\sigma=30$ & 31.29 & 0.708 & 0.1299 & 0.189 & 58.23 & 0.595 \\
Element-wise addition fusion & 35.28 & 0.839 & 0.0925 & 0.137 & 46.14 & 0.703 \\
\bottomrule[1.2pt]
\end{tabular}
}
\vspace{-10pt}
\end{table}

\noindent \textbf{Ablation on Model Architecture of Emissify3D.}
Table~\ref{tab:ablation} compares the full model with variants that isolate three design choices for adapting the video diffusion backbone to multi-view emission map generation: frame replication, temporal RoPE, and conditioning fusion. The full model performs best on all six metrics.

\noindent \emph{Frame replication.} We remove the fourfold replication of all frames except the first during both training and inference. Without replication, PSNR drops from $37.10$ to $32.48$ dB, and LPIPS rises from $0.0839$ to $0.1207$. These results support compensating for the VAE's fourfold temporal downsampling to preserve information from sparsely sampled viewpoints.

\noindent \emph{Temporal RoPE.} A stride of $\sigma=1$ recovers the backbone's original consecutive frame positions and gives the lowest PSNR ($30.76$~dB); $\sigma=30$ gives $31.29$~dB, and the full model with $\sigma=15$ gives $37.10$~dB. The stride therefore matters substantially, and an intermediate value works best for the viewpoint changes in our sequences.

\noindent \emph{Conditioning fusion.} We replace channel-wise concatenation of the reference image, emission control image, and target-view normal maps with element-wise addition of their latents, which does not require expanding the patchify layer's input channels. This variant reduces SSIM from $0.946$ to $0.839$ and CLIP-I from $0.799$ to $0.703$, favoring separate conditioning channels for both structural fidelity and alignment with the control image.

\noindent \textbf{Ablation on Emission Control Generation.}
We replace the generated emission control image with the raw user sketch. As shown in Figure~\ref{fig:control_generation_ablation}, the sketch is spatially ambiguous and poorly aligned with the object appearance, so Emissify3D produces misplaced emissive regions and incomplete structures. Grounding the guidance in the reference rendering and making emitting regions and colors explicit recovers accurate, view-consistent emission, indicating that control generation is a necessary component rather than cosmetic preprocessing.

\section{Limitations}
\label{sec:limitations}

Our approach has several limitations. First, the emission strength is a single global scalar, so assets that need different intensities for different parts (e.g., a bright headlight and dim panel trim) cannot be calibrated independently. Second, with $K=9$ views sampled around the object, regions that are poorly visible from these views (e.g., the top, bottom, or concave interiors) receive little or no supervision during back-projection. Third, the control-generation and brightness-evaluation stages rely on proprietary or very large models, which limits reproducibility and adds latency and cost; errors in the control image propagate directly to the final texture. Fourth, training data are derived from a single source dataset, and emission styles not represented in Objaverse-Emission may not be reproduced faithfully. Finally, our quantitative benchmark evaluates only multi-view emission map generation from ground-truth control images; the end-to-end quality of the pipeline from text or sketch input is demonstrated qualitatively.

\section{Conclusion}
\label{sec:conclusion}

We introduced \emph{non-emissive textured 3D asset emissification}, a preservation-oriented editing task that adds a controllable emission texture while preserving the original geometry and non-emissive PBR maps. Our pipeline translates user guidance into a reference-aligned emission control image, generates multi-view emission maps with Emissify3D, and adjusts the emission strength through VLM feedback. Emissify3D repurposes an image-to-video diffusion model through expanded input encoding, a temporal RoPE stride, and frame replication, and is trained on data built from Objaverse-Emission. Experiments show that Emissify3D generates multi-view emission maps conditioned on emission controls more faithfully than other baselines. Qualitative examples illustrate how artists can add emission without re-authoring their existing materials.

\subsubsection*{Broader Impact Statement}
This work targets content-creation tools for 3D artists. Emission editing changes only the appearance of an asset and does not create new likenesses of people, so we do not anticipate direct misuse beyond that of general image and 3D generation tools. Because the pipeline relies on proprietary image-editing and vision-language services, its outputs inherit the biases and availability constraints of those services.

\bibliography{tmlr-main}
\bibliographystyle{tmlr}

\end{document}